%% file: ms.tex
\documentclass[letterpaper, 10 pt, conference]{ieeeconf}  % Comment this line out if you need a4paper

\IEEEoverridecommandlockouts

\UseRawInputEncoding

\usepackage{graphicx} % for pdf, bitmapped graphics files
\usepackage{times} % assumes new font selection scheme installed
\usepackage{amsmath} % assumes amsmath package installed
\usepackage{amssymb}  % assumes amsmath package installed
\usepackage{url}
\usepackage{enumerate}
\usepackage{multirow}
\usepackage{colortbl}
\usepackage{booktabs} % for top and bottom line
\usepackage{algorithmicx}
\usepackage{algorithm}
\usepackage{makecell}
\usepackage{algpseudocode}% http://ctan.org/pkg/algorithmicx
\usepackage{cite}
\usepackage{bm}
\makeatletter
\let\NAT@parse\undefined
\makeatother
\usepackage[hidelinks]{hyperref} %hide the borders of the cross-references

\usepackage{xcolor}
\definecolor{gray_2}{RGB}{200, 200, 200}

\begin{document}

\title{\LARGE \bf
CMDFusion: Bidirectional Fusion Network with Cross-modality Knowledge Distillation for LIDAR Semantic Segmentation
\author{Jun Cen$^{1,2*}$, Shiwei Zhang$^{2}$, Yixuan Pei$^{3}$, Kun Li$^{2}$, Hang Zheng$^{2}$, Maochun Luo$^{2}$, Yingya Zhang$^{2}$, Qifeng Chen$^{1}$}

\thanks{$^{1}$Authors are with Cheng Kar-Shun Robotics Institute, The Hong Kong University of Science and Technology, Hong Kong SAR, China. \texttt \{{jcenaa\}}@connect.ust.hk. \{{cqf\}}@ust.hk.}
\thanks{$^{2}$Authors are with Alibaba Group, China. \texttt \{{zhangjin.zsw, zh334251, luomaochun.lmc, yingya.zyy\}}@alibaba-inc.com. \texttt \{{lk158400\}}@cainiao.com.}
\thanks{$^{3}$Authors are with the SMILES LAB at the School of Information and Communication Engineering'an Jiaotong University, Xi'an, China. \texttt \{{peiyixuan\}}@stu.xjtu.edu.}
\thanks{$^{*}$Work done as an intern at Alibaba DAMO Academy.}

}
\maketitle

%%%%%%%%%%%%%%%%%%%%%%%%%%%%%%%%%%%%%%%%%%%%%%%%%%%%%%%%%%%%%%%%%%%%%%%%%%%%%%%%
\input{latex/0_abstract.tex}
\input{latex/1_intro.tex}

\input{latex/2_related.tex}
\input{latex/3_method.tex}

\input{latex/4_exp.tex}

\input{latex/5_conclu.tex}
%%%%%%%%%%%%%%%%%%%%%%%%%%%%%%%%%%%%%%%%%%%%%%%%%%%%%%%%%%%%%%%%%%%%%%%%%%%%%%%%
\bibliographystyle{IEEEtran}
\bibliography{ref}

\end{document}

%% file: latex/0_abstract.tex
\begin{abstract}
2D RGB images and 3D LIDAR point clouds provide complementary knowledge for the perception system of autonomous vehicles. Several 2D and 3D fusion methods have been explored for the LIDAR semantic segmentation task, but they suffer from different problems. 2D-to-3D fusion methods require strictly paired data during inference, which may not be available in real-world scenarios, while 3D-to-2D fusion methods cannot explicitly make full use of the 2D information. Therefore, we propose a Bidirectional Fusion Network with Cross-Modality Knowledge Distillation (CMDFusion) in this work. Our method has two contributions. First, our bidirectional fusion scheme explicitly and implicitly enhances the 3D feature via 2D-to-3D fusion and 3D-to-2D fusion, respectively, which surpasses either one of the single fusion schemes. Second, we distillate the 2D knowledge from a 2D network (Camera branch) to a 3D network (2D knowledge branch) so that the 3D network can generate 2D information even for those points not in the FOV (field of view) of the camera. In this way, RGB images are not required during inference anymore since the 2D knowledge branch provides 2D information according to the 3D LIDAR input. We show that our CMDFusion achieves the best performance among all fusion-based methods on SemanticKITTI and nuScenes datasets. The code will be released at https://github.com/Jun-CEN/CMDFusion.
\end{abstract}

%% file: latex/1_intro.tex
\section{INTRODUCTION}

\begin{figure}[t]
\centering
\includegraphics[width=3.3in]{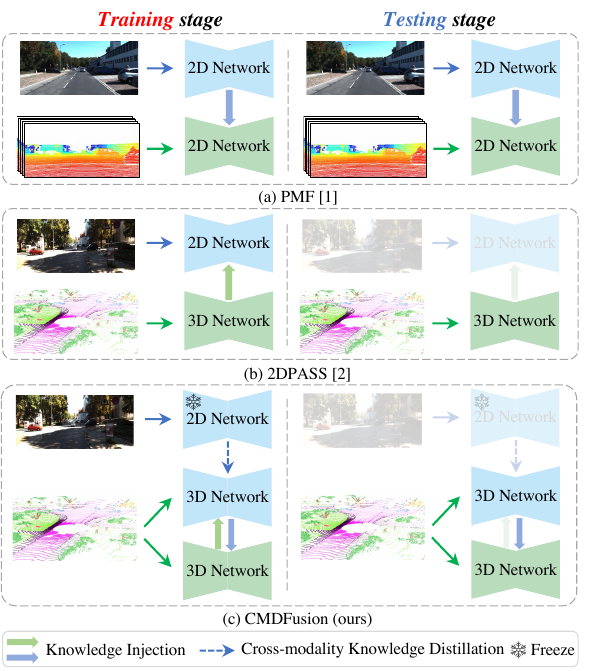}
\caption{(a) 2D-to-3D methods (PMF~\cite{perception}) can only handle points in the FOV of the camera. (b) 3D-to-2D methods (2DPASS~\cite{2dpass}) do not explicitly involve the 2D information into 3D features. (c) Our CMDFusion can process the whole point cloud through cross-modality distillation and strengthen the 3D features through bidirectional fusion.}
\label{fig:backgroud}
% \vspace{-1.5em} % for figure gap smaller
\end{figure}
% \footnotetext[3]{}

3D LIDAR is significant for the perception system of autonomous vehicles, and one of the applicable tasks with LIDAR is semantic segmentation. Great efforts have been made for better LIDAR semantic segmentation performance using single LIDAR modality~\cite{spvcnn, rangenet, cylinder3d, rpvnet}. Recently, several multi-modality methods are developed~\cite{2dpass, perception} to fuse the features of LIDAR and colorful cameras since they provide complementary information. LIDAR provides reliable depth information and is robust to light conditions such as dark nights, while the camera offers a dense colorful appearance and fine-grained textures. In this work, we also aim to study how to effectively leverage these two modality data for better LIDAR semantic segmentation.

Existing fusion-based methods can be divided into 2D-to-3D fusion method (PMF~\cite{perception}) and 3D-to-2D fusion method (2DPASS~\cite{2dpass}), as shown in Fig.~\ref{fig:backgroud} (a) and (b). PMF injects 2D knowledge into the LIDAR features, so it needs strictly paired data during training and inference. However, the FOV of LIDAR and the camera may not totally overlap with each other, so those points out of the FOV of the camera cannot be tested. For example, SemanticKITTI~\cite{semantickitti} only provides two front-view images, and points at the side and back cannot be involved in the PMF framework. 2DPASS notices this problem and proposed injecting 3D features into 2D features during training to implicitly enhance the 3D features. In this way, 2DPASS does not require images during inference. However, 3D features do not explicitly contain 2D information in such a 3D-to-2D scheme.

To solve the mentioned problems of 2D-to-3D and 3D-to-2D fusion methods, we propose a Bidirectional Fusion Network with
Cross-Modality Knowledge Distillation (CMDFusion), as shown in Fig.~\ref{fig:backgroud} (c). Specifically, on the one hand, we propose a Bidirectional Fusion Block (BFB) to explicitly and implicitly enhance the 3D features through 2D-to-3D and 3D-to-2D injection, which owns the benefits of both single fusion schemes. On the other hand, we propose a Cross-Modality Distillation (CMD) module to let a 3D network (2D knowledge branch) memorize the information of the 2D network (camera branch) during training. During inference, the 2D knowledge branch provides the 2D image information based on the 3D LIDAR point cloud inputs so that we can obtain the 2D knowledge for the whole point cloud, including those points not in the FOV of the camera.

We evaluate our method on two challenging datasets, including SemanticKITTI~\cite{semantickitti} and NuScenes~\cite{nuscenes}. Experiments show that our method achieves the best performance among all fusion-based methods. In summary, our contributions include the following:
\begin{itemize}
    \item We develop a bidirectional fusion method CMDFusion for the LIDAR semantic segmentation task, which surpasses the single directional 2D-to-3D fusion and 3D-to-2D fusion methods.
    \item We develop a cross-modality distillation module to generate 2D information for those points that are out of the FOV of the camera.
    \item We experimentally show that our method achieves the best performance among fusion-based methods on SemanticKITTI and Nuscenes datasets.
\end{itemize}

%% file: latex/2_related.tex
\section{RELATED WORK}

3D LIDAR semantic segmentation has grown very fast based on well-annotated public datasets, such as SemanticKITTI~\cite{semantickitti} and NuScenes~\cite{nuscenes}. Most methods in this area are single-modality, \textit{i.e.}, only use LIDAR point cloud to extract information. Specifically, single-modality methods can be categorized into point-based, projection-based, voxel-based, and multi-view fusion methods. 

1) Point-based methods~\cite{Randla-Net, thomas2019kpconv, wu2019pointconv} adapt PointNet~\cite{qi2017pointnet} and PointNet++~\cite{qi2017pointnet++} to the LIDAR domain. These point-based methods do not generalize very well in the LIDAR point cloud scenarios since their sampling and searching algorithms cannot perfectly handle the sparse outdoor point clouds. 

2) Voxel-based methods divide the whole point cloud into voxels~\cite{Minkowski} and apply efficient 3D convolution for semantic segmentation like SparseConv~\cite{SparseConv}. Cylinder3D~\cite{cylinder3d} proposed a cylindrical partition and asymmetrical 3D convolutional network which follows the geometry structure of the LIDAR point cloud. 

3) Projection-based methods first project 3D LIDAR point cloud into 2D range-view images~\cite{rangenet} or bird’s-eye-view (BEV) images~\cite{polarnet} and then apply 2D convolution network for semantic segmentation. However, such a projection inevitably loses some of the 3D geometry information.

4) Multi-view fusion methods combine different views of the LIDAR point cloud as inputs. FusionNet~\cite{fusionnet} and SPVCNN~\cite{spvcnn} fuse voxel and point level information, while RPVNet~\cite{xu2021rpvnet} fuses the information of voxel, point, and range views.

Recently, multi-modality fusion has become popular in the autonomous driving area. In the 3D object detection task, BEV fusion~\cite{liu2022bevfusion,liang2022bevfusion,yang2022deepinteraction} unifies the LIDAR and image features in the BEV space and achieves the state-of-the-art performance. However, the height information is much more critical in the semantic segmentation task than the object detection task, so the BEV-based method~\cite{polarnet} has limited performance on the semantic segmentation task. Instead, PMF~\cite{perception} projects the LIDAR point cloud into the image space and then conducts 2D-to-3D fusion for better 3D feature representation. 2DPASS~\cite{2dpass} finds that the 2D-to-3D fusion method like PMF can only be applied on the points in the overlapping FOVs of the LIDAR and camera, so 2DPASS conducts 3D-to-2D fusion to strengthen the 3D features by supervising the 3D features from the 2D branch. 

Compared to PMF and 2DPASS, our bidirectional fusion network enjoys the benefits of both 2D-to-3D and 3D-to-2D fusion schemes. Besides, we propose a cross-modality distillation module so that our network can be applied to the whole LIDAR point cloud, including the points that are out of the FOV of the camera.

%% file: latex/3_method.tex
\section{METHODOLOGY}

\subsection{Framework Overview}
The simplified and specific overall structure of our proposed CMDFusion is shown in Fig.~\ref{fig:backgroud} (c) and Fig.~\ref{fig:arc} (a), respectively. Our CMDFusion is composed of three branches, including a camera branch (2D network), a 2D knowledge branch (3D network), and a 3D LIDAR branch (3D network).
\subsubsection{Training}
During training, the 2D knowledge branch (a 3D network) learns the 2D image information from the camera branch (a 2D network) via Cross-Modality Distillation (CMD). Although the CMD is conducted on those points in the overlapping FOVs of the LIDAR and camera, the 2D knowledge branch can be generalized to the points that are out of the FOV of the camera. In this way, we can obtain the 2D information of the whole point cloud, which is not approachable in PMF~\cite{perception} or 2DPASS~\cite{2dpass}. Then we fuse the features of the 2D knowledge branch and 3D LIDAR branch through Bidirectional Fusion Block (BFB). On the one hand, 2D-to-3D directional fusion explicitly enhances the 3D feature via 2D information injection. On the other hand, 3D-to-2D directional fusion implicitly improves the robustness of the 3D feature since it is required to have the potential to be well adapted to the 2D space. Therefore, our BFB enjoys the benefits of both PMF and 2DPASS.

\subsubsection{Testing}
During inference, the camera branch is not needed anymore since its knowledge is already distilled to the 2D knowledge branch. Besides, only 2D-to-3D directional fusion is involved as the final prediction results come from the 3D LIDAR branch. The right-hand side of Fig.~\ref{fig:backgroud}~(c) shows the parts that are needed during inference.

\begin{figure*}[ht]
\centering
\includegraphics[width=7in]{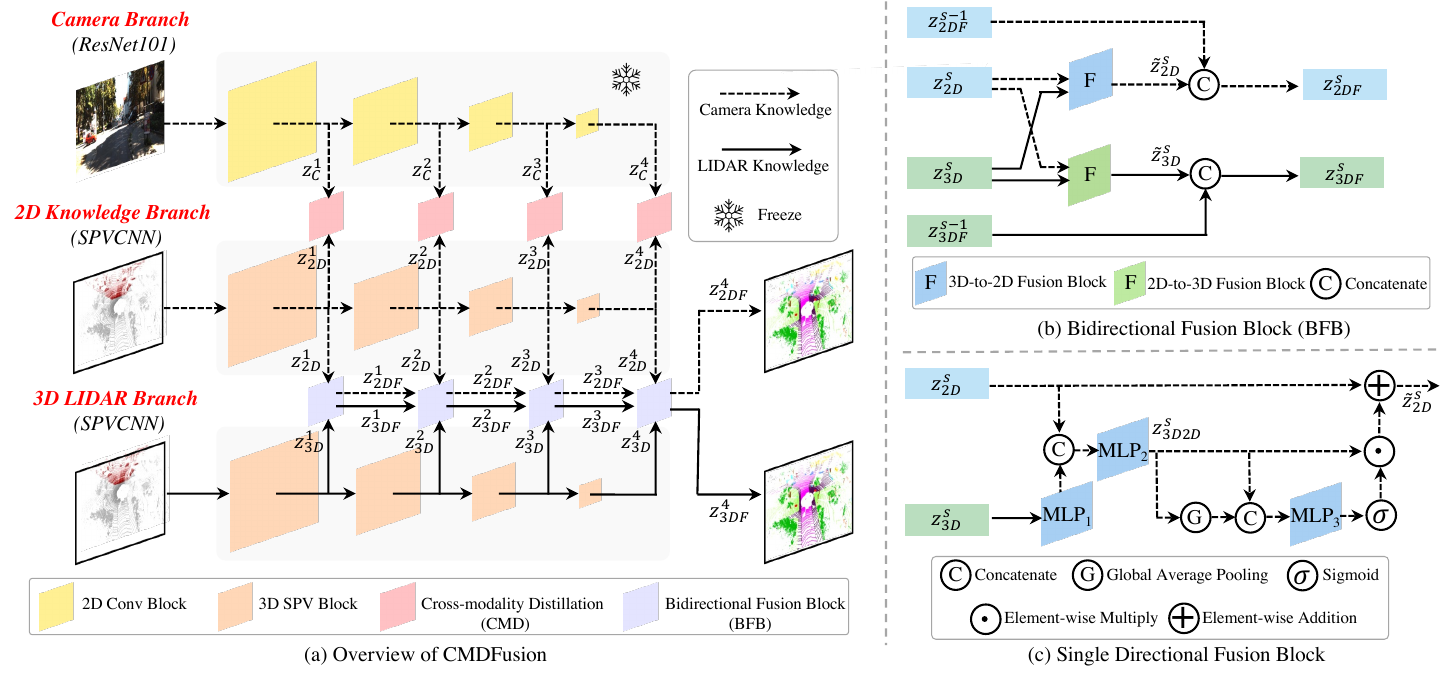}
\caption{(a) Framework overview of CMDFusion. The 2D knowledge branch learns the 2D information from the camera branch through Cross-Modality Distillation (CMD). The 3D LIDAR branch and 2D knowledge branch interact with each other through Bidirectional Fusion Block (BFB) at each scale. (b) The structure of BFB. (c) The structure of the single fusion block.}
\label{fig:arc}
% \vspace{-1.0em} % for figure gap smaller
\end{figure*}

\subsection{Point-to-pixel Corrspondence}
\label{sec:p2p}
Point-to-pixel correspondence is the pre-request of Cross-Modality Distillation (CMD). Given a LIDAR point cloud $P = \{p_i\}_{i=1}^N \in \mathbb{R}^{N\times 3}$, where $p_i = (x_i, y_i, z_i) \in \mathbb{R}^{3}$ refers to the XYZ coordinates of a point and $N$ is the number of points in the point cloud, the projected 2D coordinates of the point $p_i$ is calculated as:
\begin{equation}
    [u_i, v_i, 1]^T = \frac{1}{z_i} \times K\times T \times [x_i, y_i, z_i, 1]^T,
	\label{project}
\end{equation}
where $K \in \mathbb{R}^{3\times 4}$ and $T \in \mathbb{R}^{4\times 4}$ denote the intrinsic and extrinsic matrices of the camera, respectively. Then we have $\hat{p}_i =(\lfloor  v_i\rfloor , \lfloor  u_i \rfloor ) \in \mathbb{R}^{2}$ as the integer projected 2D coordinates, where $\lfloor \cdot \rfloor$ is the floor operation. For the SemanticKITTI dataset, $K$ and $T$ are already given. For the NuScenes dataset, the extrinsic matrix $T$ is calculated as:
\begin{equation}
    T=T_{\text {C} \leftarrow \mathrm{ego}_{\mathrm{t}_{c}}} \times 
T_{\mathrm{ego}_{\mathrm{t}_{\mathrm{c}}} \leftarrow \mathrm{G}}\times 
	T_{\mathrm{G} \leftarrow \mathrm{ego}_{\mathrm{t}_{l}}}\times 
T_{\mathrm{ego}_{\mathrm{t}_{l}} \leftarrow \mathrm{L}}
	\label{nuscene_trans},
\end{equation}
where $L$, $C$, and $G$ refer to the LIDAR, camera, and global.

Note that CMD is only applied on the points that are in the overlapping FOVs of LIDAR and camera, as shown in the colorized region in the input of the 2D knowledge branch in Fig.~\ref{fig:arc} (a). Formally, suppose the points set in the overlapping FOVs of LIDAR and camera is $P^{O} = \{p_i\}_{i=1}^{N^{O}} \in \mathbb{R}^{{N^{O}}\times 3}$, where $N^{O}$ denotes the number of points in the overlapping FOVs of the LIDAR and camera, then for each point $p_i$ in $P^{O}$, its corresponding projected coordinates $\hat{p}_i =(\lfloor  v_i\rfloor , \lfloor  u_i \rfloor )$ should meet:
\begin{equation}
\left\{
        \begin{array}{cl}
0 \le \lfloor  v_i\rfloor \le H \\
0 \le \lfloor  u_i\rfloor \le W,
        \end{array} \right.
\end{equation}
where $H$ and $W$ refer to the height and width of corresponding images. Note that for feature maps under different scales, we first upsample the feature maps to the original scale and then use the corresponding point-to-pixel corresponding.

\subsection{Cross-Modality Distillation}

Cross-Modality Distillation (CMD) is to distillate the 2D knowledge from the camera branch (a 2D network) to the 2D knowledge branch (a 3D network), so we can generate the 2D information for those points out of the FOV of the camera and do not need the images during inference. 
\subsubsection{Camera Branch}
Unlike PMF~\cite{perception} and 2DPASS~\cite{2dpass} that train the camera branch with the ground truth projected from the LIDAR point cloud, we use a ResNet101~\cite{he2016deep} which is pre-trained on the Cityscapes dataset~\cite{cordts2016cityscapes}. Cityscapes is a popular dataset for 2D image semantic segmentation in the autonomous driving scenario. We adopt this strategy for two reasons. First, if we use the ground truth which is projected from the LIDAR point cloud, the camera branch may learn the overlapping knowledge with the 3D LIDAR branch since they share the same ground truth source. In contrast, the pre-trained camera branch using another dataset could provide additional information on top of the LIDAR point cloud. Second, we could freeze the camera branch during training since it is well-trained, so less back-propagation is needed for the whole structure. In this way, the training process consumes less GPU memory and time.

\subsubsection{2D Knowledge Branch} Following 2DPASS~\cite{2dpass}, we use SPVCNN~\cite{spvcnn} as the 3D network used in this paper, including the 2D knowledge branch and 3D LIDAR branch. Now let us formulate the process of CMD. 

For points in the overlapping FOVs of LIDAR and camera $p_i \in P^O$, we feed them into the 2D knowledge branch $f_{2D}$ to obtain the features $z_{2D}^s$:
\begin{equation}
    z_{2D}^s=\{ f_{2D}^{s}(p_i) \}_{i=1}^{N^{O}} \in \mathbb{R}^{{N^{O}}\times d}, 
\end{equation}
where $s=\{1,2,3,4 \}$ and $d$ refer to the feature map scale and the dimension of the features, respectively. Then we obtain the corresponding features $z_C^s$ of $P^O$ from the camera branch through the point-to-pixel projection described in Sec.~\ref{sec:p2p}. The CMD is realized through this loss $\mathcal{L}_{CMD}$:
\begin{equation}
    \mathcal{L}_{CMD} = \frac{1}{N^O}\sum \left \| z_{2D}^s - z_{C}^s  \right \|_2,
\end{equation}
where $\left \| \cdot  \right \|_2$ denotes the L2 loss. In this way, the 2D knowledge branch can mimic the function of the camera branch to provide the 2D information based on the 3D LIDAR point cloud. Although $\mathcal{L}_{CMD}$ is only available for $P^O$ during training, the trained 2D knowledge branch can be generalized to the whole point cloud $P$ during inference.

\subsection{Bidirectional Fusion}
Our bidirectional fusion block (BFB) is composed of a 3D-to-2D fusion block and a 2D-to-3D fusion block, as shown in Fig.~\ref{fig:arc} (b). 2D-to-3D directional fusion explicitly enhances the 3D features via 2D feature injection, while 3D-to-2D implicitly enhances the 3D features via 2D knowledge branch supervision. Note that the 3D-to-2D fusion block and 2D-to-3D fusion block share the same single directional fusion structure, as shown in Fig.~\ref{fig:arc} (c), and the only difference is the input position. Fig.~\ref{fig:arc} (c) is the example of the 3D-to-2D single directional fusion block, and we can obtain the 2D-to-3D single directional fusion block by simply changing the positions of two inputs in Fig.~\ref{fig:arc} (c). Unlike CMD which can only be applied on the $P^{O}$, BFB is applied on the whole point cloud. So $z_{2D}^s \in \mathbb{R}^{N\times d}$ and $z_{3D}^s\in \mathbb{R}^{N\times d}$ in this section.
\subsubsection{3D-to-2D Fusion}
3D-to-2D fusion is illustrated in Fig.~\ref{fig:arc} (c). Formally, we first have:
\begin{equation}
    z_{3D2D}^s = \texttt{MLP}_2(\texttt{Cat}(\texttt{MLP}_1(z_{3D}^s), z_{2D}^s)),
\end{equation}
where $\texttt{MLP}$ is a multiplayer perceptron, and $\texttt{Cat}$ refers to the feature concatenation. $\texttt{MLP}_1$ is used to transfer the 3D feature $z_{3D}^s$ into the 2D feature space. $\texttt{MLP}_2$ is responsible to transfer the concatenated feature into the residual space of $z_{2D}^s$. Then we have:
\begin{small}
\begin{equation}
\tilde{z}_{2D}^s = z_{2D}^s \oplus \sigma(\texttt{MLP}_3(\texttt{Cat}(\texttt{GAP}(z_{3D2D}^s),z_{3D2D}^s))) \odot z_{3D2D}^s,
\end{equation}
\end{small}
where $\oplus$ and $\odot$ denote the element-wise plus and element-wise multiply, respectively. $\texttt{GAP}$ means global average pooling, and $\sigma$ means Sigmoid activation function. $\texttt{GAP}$ is used to integrate the gloable information, and $\texttt{MLP}_3$ is used to transfer the feature into the attention value. $\tilde{z}_{2D}^s$ represents the enhanced 2D features of scale $s$. Then we concatenate $\tilde{z}_{2D}^s$ and the enhanced features of previous scales $z_{2DF}^{s-1}$ to obtain $z_{2DF}^{s}$:
\begin{equation}
    z_{2DF}^{s} = \texttt{Cat}(z_{2DF}^{s-1},\tilde{z}_{2D}^s),
\end{equation}
where $z_{2DF}^{s}$ contains all enhanced 2D features from scale 1 to $s$. Finally, $z_{2DF}^{4}$ contains the enhanced 2D features of all 4 scales, and we use a linear classifier $g_{2D}$ to output the logits. The loss of 2D knowledge branch $\mathcal{L}_{2D}$ is formulated as:
\begin{equation}
    \mathcal{L}_{2D} = -\frac{1}{N}\sum ylog(g_{2D}(z_{2DF}^{4})_{y}),
\end{equation}
where $y$ refers to the ground truth, and $g(z_{2DF}^{4})_{y}$ denotes the $y^{\text{th}}$ logit of $g(z_{2DF}^{4})$. Note that single directional fusion does not share MLPs for different scales.

\subsubsection{2D-to-3D Fusion} 2D-to-3D fusion shares the symmetric structure with 2D-to-3D fusion. Formally, we have the following:
\begin{small}
\begin{equation}
 \begin{aligned}
 z_{2D3D}^s = \texttt{MLP}_2(\texttt{Cat}&(\texttt{MLP}_1(z_{2D}^s), z_{3D}^s)),  \\
 \tilde{z}_{3D}^s = z_{3D}^s \oplus \sigma(\texttt{MLP}_3(\texttt{Cat}(\texttt{GAP}&(z_{2D3D}^s),z_{2D3D}^s))) \odot z_{2D3D}^s,  \\
  z_{3DF}^{s} = \texttt{Cat}(&z_{3DF}^{s-1},\tilde{z}_{3D}^s).  \\
  \end{aligned}
 \end{equation}
 \end{small}
 Similarly, $z_{3DF}^{4}$ is the final enhanced 3D feature, and a linear classifier $g_{3D}$ is used to output the logits. The loss of 3D knowledge branch $\mathcal{L}_{3D}$ is formulated as:
\begin{equation}
    \mathcal{L}_{3D} = -\frac{1}{N}\sum ylog(g_{3D}(z_{3DF}^{4})_{y}).
\end{equation}
Note that 2D-to-3D fusion blocks do not share MLPs and classifiers with 3D-to-2D fusion blocks.

\subsection{Overall Training and Testing Process}
\subsubsection{Training} The overall loss $\mathcal{L}_{all}$ for training the model is calculated as:
\begin{equation}
    \mathcal{L}_{all} =  \mathcal{L}_{CMD} +  \mathcal{L}_{2D} +  \mathcal{L}_{3D}.
\end{equation}
\subsubsection{Testing} We use the output of the classifier in the 3D LIDAR branch as the final prediction results. Specifically, the prediction result $\hat{y}$ is:
\begin{equation}
    \hat{y} = \mathop{\arg\max}\limits_{i=1,2,...,C} \ g_{3D}(z_{3DF}^{4})_{i},
\end{equation}
where $C$ denotes the total number of classes in the dataset.

%% file: latex/4_exp.tex
\section{EXPERIMENTS}

\begin{table*}
\centering
\caption{Comparisons on \textit{SemanticKITTI-O} validation set. \textit{L} and \textit{C} refer to LIDAR and camera modality.}
\small
\renewcommand\tabcolsep{4pt}
% \vskip 0.15in
 % scalebox{0.6}
% \renewcommand\arraystretch{1.1}
\resizebox{\textwidth}{!}{
\begin{tabular}{l|cc|c|ccccccccccccccccccc}
\hline
Method & Train & Test &\rotatebox{90}{mIoU} & \rotatebox{90}{car} & \rotatebox{90}{bicycle} & \rotatebox{90}{motorcycle} & \rotatebox{90}{truck} & \rotatebox{90}{other-vehicle} & \rotatebox{90}{person} & \rotatebox{90}{bicyclist} & \rotatebox{90}{motorcyclist} & \rotatebox{90}{road} & \rotatebox{90}{parking} & \rotatebox{90}{sidewalk} & \rotatebox{90}{other-ground} & \rotatebox{90}{building} & \rotatebox{90}{fence} & \rotatebox{90}{vegetation} & \rotatebox{90}{trunk} & \rotatebox{90}{terrain} & \rotatebox{90}{pole} & \rotatebox{90}{traffic-sign} \\ %  & \rotatebox{90}{fwIoU(\%)} \\

\hline\hline
% \#Points (\textit{k}) & - & 6384 & 44 & 52 & 101 & 471 & 127 & 129 & 5 & 21434 & 974 & 8149 & 67 & 6304 & 1691 & 20391 & 882 & 8125 & 317 & 64 & -\\
% \hline
RandLANet~\cite{Randla-Net} & L & L & 50.0 
& 92.0 & 8.0 & 12.8 & {74.8} & 46.7 & 52.3 & 46.0 & 0.0 & 93.4 & 32.7 & 73.4 & 0.1 & 84.0 & 43.5 & 83.7 & 57.3 & {73.1} & 48.0 & 27.3 \\ 
RangeNet++~\cite{rangenet} & L &L & 51.2 
& 89.4 & 26.5 & 48.4 & 33.9 & 26.7 & 54.8 & 69.4 & 0.0 & 92.9 & 37.0 & 69.9 & 0.0 & 83.4 & 51.0 & 83.3 & 54.0 & 68.1 & 49.8 & 34.0 \\ %  & 81.0 \\
SequeezeSegV2~\cite{wu2019squeezesegv2} & L &L & 40.8 
& 82.7 & 15.1 & 22.7 & 25.6 & 26.9 & 22.9 & 44.5 & 0.0 & 92.7 & 39.7 & 70.7 & 0.1 & 71.6 & 37.0 & 74.6 & 35.8 & 68.1 & 21.8 & 22.2 \\ % & 76.4 \\
SequeezeSegV3~\cite{xu2020squeezesegv3} & L &L & 53.3 
& 87.1 & 34.3 & 48.6 & 47.5 & 47.1 & 58.1 & 53.8 & 0.0 & {95.3} & 43.1 & {78.2} & 0.3 & 78.9 & 53.2 & 82.3 & 55.5 & 70.4 & 46.3 & 33.2 \\ % & 82.2 \\

SalsaNext~\cite{cortinhal2020salsanext} & L &L  & 59.4 
& 90.5 & 44.6 & 49.6 & 86.3 & 54.6 & 74.0 & 81.4 & 0.0 & 93.4 & 40.6 & 69.1 & 0.0 & 84.6 & 53.0 & 83.6 & 64.3 & 64.2 & 54.4 & 39.8\\ % & 81.9 \\

% ~\cite{4D Spatio-Temporal ConvNets: Minkowski Convolutional Neural Networks. }
MinkowskiNet~\cite{choy20194d} & L &L & 58.5 
& 95.0 & 23.9 & 50.4 & 55.3 & 45.9 & 65.6 & 82.2 & 0.0 & 94.3 & 43.7 & 76.4 & 0.0 & 87.9 & 57.6 & 87.4 & 67.7 & 71.5 & {63.5} & {43.6} \\ % & 85.1 \\

% spvnas
SPVNAS~\cite{spvcnn} & L &L & 62.3
& 96.5 & 44.8 & {63.1} & 59.9 & 64.3 & 72.0 & {86.0} & 0.0 & 93.9 & 42.4 & 75.9 & 0.0 & 88.8 & {59.1} & {88.0} & 67.5 & 73.0 & {63.5} & 44.3  \\ % & 85.6 \\

Cylinder3D~\cite{cylinder3d} & L  &L & 64.9
& {96.4} & \textbf{61.5} & 78.2 & 66.3 & {69.8} & 80.8 & \textbf{93.3} & 0.0 & 94.9 & 41.5 & 78.0 & \textbf{1.4} & 87.5 & 50.0 & 86.7 & {72.2} & 68.8 & 63.0 & 42.1 \\ \hline%  & 85.1 \\ 
PointPainting~\cite{vora2020pointpainting} & L+C &L+C & 54.5
& 94.7 & 17.7 & 35.0 & 28.8 & 55.0 & 59.4 & 63.6 & 0.0 & {95.3} & 39.9 & 77.6 & 0.4 & 87.5 & 55.1 & 87.7 & 67.0 & 72.9 & 61.8 & 36.5 \\

% FuseSeg*~\cite{krispel2020fuseseg} & L+C \\ 
RGBAL~\cite{Madawy2019RGBAL} & L+C &L+C & 56.2 
& 87.3 & 36.1 & 26.4 & 64.6 & 54.6 & 58.1 & 72.7 & 0.0 & 95.1 & 45.6 & 77.5 & {0.8} & 78.9 & 53.4 & 84.3 & 61.7 & 72.9 & 56.1 & 41.5 \\
PMF~\cite{perception} & L+C &L+C & {63.9}
& 95.4 & {47.8} & 62.9 & 68.4 & \textbf{75.2} & {78.9} & 71.6 & 0.0 & \textbf{96.4} & {43.5} & 80.5 & 0.1 & {88.7} & 60.1 & 88.6 & \textbf{72.7} & 75.3 & 65.5 & 43.0 \\
% CMDFusion (Ours) & L+C &L &69.3 &96.3 &64.0 &82.7 &85.8 &75.0 &86.0 &95.2 &0.0 &96.2 &45.8 &80.0 &0.1 &89.8 &65.1 &89.7 &74.9 &76.5 &64.2 &49.6\\
\bf{CMDFusion (Ours)} & L+C &L &\textbf{70.1} &\textbf{96.9} &57.2 &\textbf{86.8} &\textbf{96.0} &69.0 &\textbf{82.7} &91.8 &\textbf{0.1} &94.8 &\textbf{51.4} &\textbf{83.1} &0.3 &\textbf{92.4} &\textbf{69.6} &\textbf{89.5} &72.2 &\textbf{77.0} &\textbf{66.9} &\textbf{53.9}\\
\hline
\end{tabular}
% \end{center}
}
\label{tab:semanti_kitti_o}
\end{table*}

\begin{figure*}[ht]
\centering
\includegraphics[width=7in]{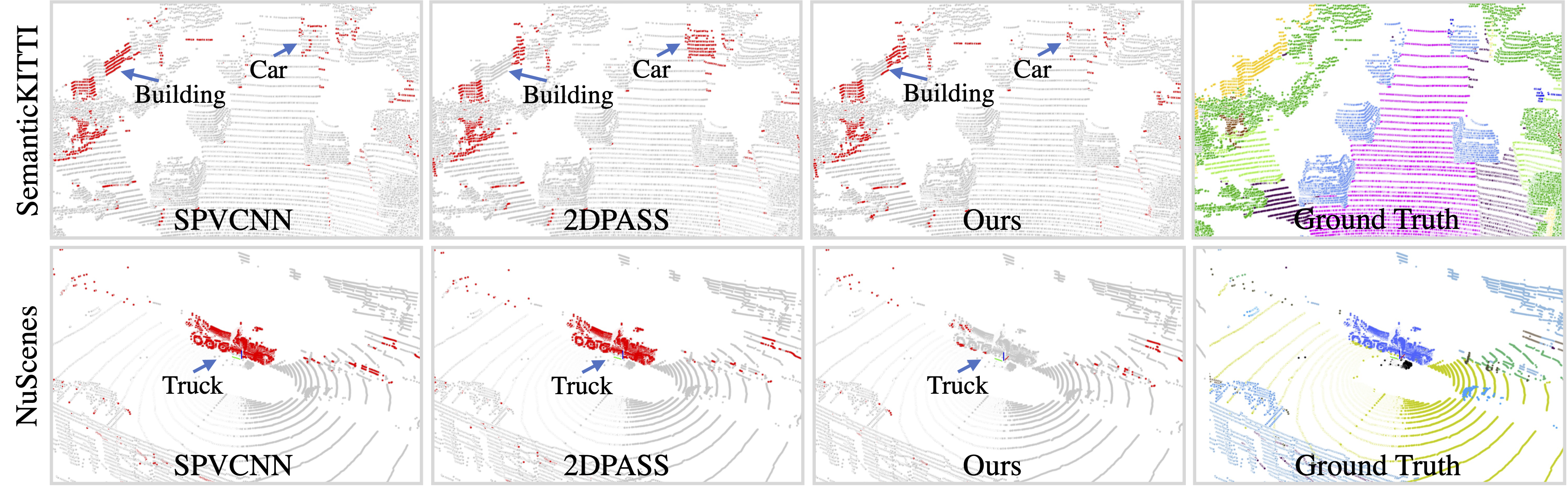}
\vspace{-0.8cm}
\caption{Error visualization samples from SemanticKITTI and NuScenes datasets. Errors are in red color. We provide the results of SPCVNN (a 3D network), 2DPASS (a 3D-to-2D fusion network), and our method (a bidirectional fusion network).}
\label{fig:vis}
\vspace{-1.0em} % for figure gap smaller
\end{figure*}

\subsection{Experiment Settings}
\subsubsection{Datasets} We conduct experiments on three large-sclae outdoor datasets, including SemanticKITTI~\cite{semantickitti}, SemanticKITTI-O~\cite{perception} and Nuscenes~\cite{nuscenes}. SemanticKITTI provides the dense segmentation labels for 00-10 sequences, in which sequence 08 is used for validation and others are used for training. The ground truth of sequences 11-21 is not reachable to the public and is used for testing. Two front-view colorful images are equipped with each LIDAR scan in SemanicKITTI. We use the image captured by the left camera in our experiments. NuScenes contains 8130 samples for training, 6019 samples for validation, and 6008 samples for testing. Six images are equipped for every LIDAR scan in Nuscenes, and we randomly pick up one image for training. SemanicKITTI-O is a subset of SemanticKITTI, which contains the points in the overlapping FOVs of the camera and LIDAR. The reason that PMF~\cite{perception} proposed the SemanicKITTI-O is that PMF cannot be applied on the points that are out of the FOV of the camera because of its 2D-to-3D fusion scheme.

\subsubsection{Evaluation Metrics}
We adopt the commonly used mean intersection-over-union (mIoU) of all classes as the evaluation metric. Specifically, mIoU is formulated as:
\begin{equation}
    mIoU = \frac{TP_c}{TP_c + FP_c + NP_c}.
\end{equation}
In addition, we also report the frequency-weighted IOU (fwIoU) provided by the NuScenes leaderboard. FwIoU is a weighted version of mIoU by the point-level frequency of different classes.

\subsubsection{Network Settings}
The camera branch is a ResNet101~\cite{he2016deep} network pre-trained using Cityscpaes~\cite{cordts2016cityscapes} dataset. Following 2DPASS~\cite{2dpass}, the 2D knowledge branch and 3D LIDAR branch are two modified SPVCNN~\cite{spvcnn} with the same structure. The feature maps from three branches are firstly reduced to the dimension of 128 and 256 for SemanticKITTI and NuScenes datasets, and then they are upsampled through bilinear interpolation to the original scale and used for CMD and BFB. As shown in Fig.~\ref{fig:arc} (a), we use feature maps from 4 scales for better performance.

\subsubsection{Training and Inference Details} Our model is trained in an end-to-end manner with the SGD optimizer. The initial learning rate is set to be 0.24, following 2DPASS~\cite{2dpass} and SPVCNN~\cite{spvcnn}. We train the model for 128 epochs for SemanticKITTI and 80 epochs for NuScenes dataset. We use the commonly used augmentation strategy in the LIDAR semantic segmentation, including global scaling with a random scaling factor sampled from [0.95, 1.05], and global rotation around the Z axis with a random angle. Image augmentation includes horizontal flipping and color jitter. The cropped image size is 1200 $\times$ 360 ($W \times H$) for SemanticKITTI and 400 $\times$ 240 for NuScenes. The voxel size in the 2D knowledge branch and 3D LIDAR branch is set to 0.1. We train our model with batch size 8 on 2 Nvidia Tesla A100 GPUs with 80G memory.

\subsection{Results on Benchmarks}

\begin{table*}[t]
\small 
\caption{Semantic segmentation results on the \textit{SemanticKITTI} test benchmark. ${\dagger}$ indicates reproduced using the official codebase. ${\ddagger}$ indicates using our implemented instance-level data augmentation.}
\vspace{-0.8cm}
\renewcommand\tabcolsep{4pt}
\begin{center}
\resizebox{\textwidth}{!}{
\begin{tabular}{l|cc|c|cccccccccccccccccccc}
\hline
                    
Method& Train & Test &
\rotatebox{90}{mIoU}&
\rotatebox{90}{road}&
\rotatebox{90}{sidewalk}&
\rotatebox{90}{parking}&
\rotatebox{90}{other-ground~}&
\rotatebox{90}{building}&
\rotatebox{90}{car}&
\rotatebox{90}{truck}&
\rotatebox{90}{bicycle}&
\rotatebox{90}{motorcycle}&
\rotatebox{90}{other-vehicle}&
\rotatebox{90}{vegetation}&
\rotatebox{90}{trunk}&
\rotatebox{90}{terrain}&
\rotatebox{90}{person}&
\rotatebox{90}{bicyclist}&
\rotatebox{90}{motorcyclist}&
\rotatebox{90}{fence}&
\rotatebox{90}{pole}&
\rotatebox{90}{traffic sign}\\
\hline
\hline

SqueezeSegV2~\cite{wu2019squeezesegv2}  &L &L &39.7&88.6& 67.6& 45.8& 17.7& 73.7& 81.8& 13.4& 18.5& 17.9& 14.0& 71.8& 35.8 &60.2& 20.1& 25.1& 3.9& 41.1& 20.2& 26.3\\

DarkNet53Seg~\cite{semantickitti} &L &L &49.9 &91.8 &74.6 &64.8 &27.9 &84.1 &86.4 &25.5 &24.5 &32.7 &22.6 &78.3 &50.1 &64.0 &36.2 &33.6 &4.7 &55.0 &38.9 &52.2\\

RangeNet53++~\cite{rangenet} &L &L &52.2 &91.8 &75.2 &65.0 &27.8 &87.4 &91.4 &25.7 &25.7 &34.4 &23.0 &80.5 &55.1 &64.6 &38.3 &38.8 &4.8 &58.6 &47.9 &55.9\\

3D-MiniNet~\cite{alonso20203d} &L &L &55.8 &91.6 &74.5 &64.2 &25.4 &89.4 &90.5 &28.5 &42.3 &42.1 &29.4 &82.8 &60.8 &66.7 &47.8 &44.1 &14.5 &60.8 &48.0 &56.6\\
SqueezeSegV3~\cite{xu2020squeezesegv3} &L &L &55.9 &91.7 &74.8 &63.4 &26.4 &89.0 &92.5 &29.6 &38.7 &36.5 &33.0 &82.0 &58.7 &65.4 &45.6 &46.2 &20.1 &59.4 &49.6 &58.9\\

PointNet++~\cite{qi2017pointnet++}&L &L &20.1& 72.0 &41.8 &18.7& 5.6 &62.3 &53.7 &0.9 &1.9 &0.2& 0.2& 46.5 &13.8 &30.0 &0.9 &1.0 &0.0 &16.9 &6.0 &8.9\\
TangentConv~\cite{tatarchenko2018tangent} &L &L &40.9 &83.9 &63.9 &{33.4} &{15.4} &{83.4} &{90.8} &15.2&{2.7}& 16.5 &12.1 &79.5 &49.3 &58.1 &23.0 &28.4 &{8.1} &{49.0} &35.8 &28.5\\
PointASNL~\cite{yan2020pointasnl}  &L &L & 46.8&87.4 &74.3&24.3&1.8&83.1&87.9&39.0&0.0&25.1&29.2&84.1&52.2&70.6&34.2& 57.6&0.0&43.9&57.8&36.9\\
RandLA-Net~\cite{Randla-Net} &L &L &55.9 &90.5 &74.0 &61.8 &24.5 &89.7 &94.2 &43.9 &29.8 &32.2 &{39.1} &83.8 &63.6 &68.6 &48.4 &47.4 &9.4 &60.4 &51.0 &50.7\\
KPConv~\cite{thomas2019kpconv} &L &L &58.8 &{90.3} &72.7 &{61.3} &31.5 &90.5 &95.0 &33.4 &30.2 &42.5 &44.3 &84.8 &69.2 &69.1 &61.5 &61.6 &11.8 &64.2 &56.4 &47.4\\

PolarNet~\cite{polarnet}&L &L &54.3 &90.8 &74.4 &61.7 &21.7 &90.0 &93.8 &22.9 &40.3 &30.1 &28.5 &84.0 &65.5 &67.8 &43.2 &40.2 &5.6 &61.3 &51.8 &57.5\\

JS3C-Net~\cite{yan2020pointasnl}   &L &L & 66.0 & 88.9& 72.1& 61.9& 31.9& 92.5& 95.8& 54.3 & 59.3& 52.9& 46.0& 84.5 & 69.8 & 67.9& 69.5 & 65.4 & 39.9 & 70.8 & 60.7 & 68.7\\
SPVNAS~\cite{spvcnn}  &L &L & 67.0  & 90.2  & 75.4  & 67.6  & 21.8  & 91.6  & 97.2  & 56.6  & 50.6  & 50.4  & 58.0  & 86.1  & 73.4  & 71.0  & 67.4  & 67.1  & 50.3  & 66.9  & 64.3  & 67.3 \\
Cylinder3D~\cite{cylinder3d}  &L &L & 68.9  & 92.2  & 77.0  & 65.0  & 32.3  & 90.7  & 97.1  & 50.8  & 67.6  & 63.8  & 58.5  & 85.6  & 72.5  & 69.8  & 73.7  & 69.2  & 48.0  & 66.5  & 62.4  & 66.2\\
RPVNet~\cite{rpvnet}  &L &L & 70.3  & \bf{93.4}  & \bf{80.7}  & \bf{70.3}  & 33.3  & \bf{93.5}  & \bf{97.6}  & 44.2  & \bf{68.4}  & 68.7  & 61.1  & \bf{86.5}  & \bf{75.1}  & \bf{71.7}  & 75.9  & 74.4  & 43.4  & \bf{72.1}  & \bf{64.8}  & 61.4 \\
(AF)$^2$-S3Net~\cite{cheng20212}  &L &L & 70.8  & 92.0  & 76.2  & 66.8  & \textbf{45.8}  & 92.5  & 94.3  & 40.2  & 63.0  & \bf{81.4}  & 40.0  & 78.6  & 68.0  & 63.1  & 76.4  & \bf{81.7}  & \bf{77.7}  & 69.6  & 64.0  & \bf{73.3}\\ \hline
SPVCNN~\cite{spvcnn} &L &L &66.2 &90.0 &74.0 &60.1 &27.3 &91.9 &96.6 &52.9 &51.4 &58.0 &58.0 &85.4 &72.0 &69.3 &\bf{77.9} &80.9 &15.8 &67.5 &60.4 &68.2\\
% \cj {2DPASS~\cite{2dpass}}  &\cj {L+C}&\cj {L}& \cj {{72.9}} & \cj{ 89.7} & \cj{ 74.7} & \cj{ 67.4} & \cj{ 40.0} & \cj{ {93.5}} & \cj{ 97.0} & \cj{ {61.1}} & \cj{ 63.6} & \cj{ 63.4} & \cj{ \bf{61.5}} & \cj{ 86.2} & \cj{ 73.9} & \cj{ 71.0} & \cj{ \bf{77.9}} & \cj{ 81.3} & \cj{ 74.1} & \cj{ \bf{72.9}} & \cj{ \bf{65.0}} & \cj{70.4}\\
2DPASS$^{\dagger}$~\cite{2dpass}  &L+C &L &67.7 &90.1 &75.1 &62.5 &30.4 &91.2 &96.6 &\bf{55.1} &61.8 &60.6 &60.0 &86.1 &72.4 &70.4 &75.3 &79.4 &22.2 &65.6 &63.3 &68.4 \\
\textbf{CMDFusion (ours)} &L+C &L &68.6 &90.0 &74.4 &64.7 &32.3 &91.5 &96.5 &52.0 &57.8 &52.5 &52.1 &85.9 &72.7 &70.7 &71.9 &80.0 &60.5 &68.0 &63.3 &67.3\\ \hline
2DPASS$^{\ddagger}$~\cite{2dpass}  &L+C &L &71.0 &89.7 &74.9 &66.8 &33.1 &92.0 &96.8 &53.1 &59.4 &66.6 &\bf{63.5} &85.8 &74.0 &70.5 &76.8 &79.7 &65.2 &67.4 &62.4 &70.6 \\
\textbf{CMDFusion$^{\ddagger}$ (ours)} &L+C &L &\bf{71.6} &90.1 &75.3 &66.9 &30.7 &91.5 &96.8 &53.1 &63.0 & 67.4 &61.2 &85.5 &74.4 &69.9 &76.6 &81.5 &77.2 &68.0 &64.5 &67.5 \\
\hline
\end{tabular} 
}
\end{center}
\label{tab:kitti_seg}
\end{table*}

\begin{table*}[t]
% \small
\renewcommand\tabcolsep{1.5pt} 
\caption{Semantic segmentation results on the \textit{Nuscenes} test benchmark. ${\dagger}$ indicates indicates reproduced using the official codebase.}
\vspace{-0.8cm}
\begin{center}
\small
\renewcommand\tabcolsep{4pt}
\resizebox{\textwidth}{!}{%
\begin{tabular}{l|cc|cc|cccccccccccccccc}
\hline

Method& Train &Test&
\rotatebox{90}{mIoU}&
\rotatebox{90}{fw mIoU}&
\rotatebox{90}{barrier}&
\rotatebox{90}{bicycle}&
\rotatebox{90}{bus}&
\rotatebox{90}{car}&
\rotatebox{90}{construction}&
\rotatebox{90}{motorcycle}&
\rotatebox{90}{pedestrian}&
\rotatebox{90}{traffic cone}&
\rotatebox{90}{trailer}&
\rotatebox{90}{truck}&
\rotatebox{90}{driveable}&
\rotatebox{90}{other flat}&
\rotatebox{90}{sidewalk}&
\rotatebox{90}{terrain}&
\rotatebox{90}{manmade}&
\rotatebox{90}{vegetation}\\
\hline
\hline

\text{PolarNet}~\cite{polarnet} & L & L& 69.4 & 87.4 & 72.2 & 16.8 & 77.0 & 86.5 & 51.1 & 69.7 & 64.8 & 54.1 & 69.7 & 63.5 & 96.6 & 67.1 & 77.7 & 72.1 & 87.1 & 84.5 \\
\text{JS3C-Net}~\cite{yan2020pointasnl} & L & L & 73.6 & 88.1 & 80.1 & 26.2 & 87.8 & 84.5 & 55.2 & 72.6 & 71.3 & 66.3 & 76.8 & 71.2 & 96.8 & 64.5 & 76.9 & 74.1 & 87.5 & 86.1 \\
\text{Cylinder3D}~\cite{cylinder3d} & L & L & 77.2 & 89.9 & 82.8 & 29.8 & 84.3 & 89.4 & 63.0 & 79.3 & 77.2 & 73.4 & 84.6 & 69.1 & \bf{97.7} & \bf{70.2} & \bf{80.3} & 75.5 & 90.4 & 87.6  \\
\text{AMVNet}~\cite{liong2020amvnet} & L & L & 77.3 & 90.1 & 80.6 & 32.0 & 81.7 & 88.9 & 67.1 & 84.3 & 76.1 & 73.5 & 84.9 & 67.3 & 97.5 & 67.4 & 79.4 & 75.5 & 91.5 & 88.7\\
\text{SPVCNN}~\cite{spvcnn} & L & L &77.4 & 89.7 & 80.0 & 30.0 & 91.9 & 90.8 & 64.7 & 79.0 & 75.6 & 70.9 & 81.0 & 74.6 & 97.4 & 69.2 & 80.0 & 76.1 & 89.3 & 87.1\\
\text{(AF)$^2$-S3Net}~\cite{cheng20212} & L & L & 78.3 & 88.5 & 78.9 & 52.2 & 89.9 & 84.2 & \bf{77.4} & 74.3 & 77.3 & 72.0 & 83.9 & 73.8 & 97.1 & 66.5 & 77.5 & 74.0 & 87.7 & 86.8\\ \hline
\text {PMF}~\cite{perception}  & L+C & L+C &77.0 & 89.0 & 82.0 & 40.0 & 81.0 & 88.0 & 64.0 & 79.0 & 80.0 & \bf{76.0} & 81.0 & 67.0 & 97.0 & 68.0 & 78.0 & 74.0 & 90.0 & 88.0\\
\text {2D3DNet}~\cite{genova2021learning} & L+C & L+C &80.0 & 90.1 & 83.0 & \bf{59.4} & 88.0 & 85.1 & 63.7 & 84.4 & \bf{82.0} & \bf{76.0} & 84.8 & 71.9 & 96.9 & 67.4 & 79.8 & \bf{76.0} & \bf{92.1} & \bf{89.2}\\
% \cj{\text {2DPASS}~\cite{2dpass}} & \cj{L+C} & \cj{L} &\cj{80.8} & \cj{90.1} & \cj{81.7} & \cj{55.3} & \cj{92.0} & \cj{91.8} & \cj{73.3} & \cj{86.5} & \cj{78.5} & \cj{72.5} & \cj{84.7} & \cj{75.5} & \cj{97.6} & \cj{69.1} & \cj{79.9} & \cj{75.5} & \cj{90.2} & \cj{88.0} \\
\text {2DPASS}$^{\dagger}$~\cite{2dpass} & L+C & L &77.8 & 89.7 & 82.4 & 31.2 & 93.6 & 91.3 & 62.6 & 81.4 & 75.4 & 69.2 & 84.3 & 76.6 & 97.3 & 68.2 & 79.5 & 75.2 & 89.3 & 87.4 \\
\textbf {CMDFusion (Ours)} & L+C & L &\bf{80.8} &\bf{90.3} & \bf{83.5} & 45.7 & \bf{94.5} & \bf{91.4} & 76.7 & \bf{87.0} & 77.2 & 73.0 & \bf{85.6} &\bf{77.3} & 97.4 & 69.2 & 79.5 & 75.5 & 91.0 & 88.5\\ \hline
\end{tabular} }
\end{center}
\label{tab:nus_seg}
\end{table*}

\subsubsection{Results on SemanticKITTI-O} PMF~\cite{perception} provides the comprehensive benchmark on the SemanticKITTI-O validation set, as shown in Table~\ref{tab:semanti_kitti_o}. The traditional 2D-to-3D fusion methods like PointPainting~\cite{vora2020pointpainting}, RGBAL~\cite{Madawy2019RGBAL}, and PMF conduct both training and inference based on the LIDAR and camera modality data, while our CMDFusion is trained on the LIDAR and camera pairs, but does not require the camera data during inference. We can see that our method significantly surpasses the PMF method by 6.2 mIoU. Note that our CMDFusion can be trained on the whole SemanticKITTI dataset based on our 2D knowledge branch and CMD, while PointPainting, RGBAL, and PMF can be only trained on the training set of SemanticKITTI-O due to their 2D-to-3D fusion scheme.

\subsubsection{Results on SemanticKITTI} Similar to 2DPASS~\cite{2dpass}, our CMDFusion is trained on the LIDAR and camera modality, while only LIDAR modality is required during inference, so 2DPASS and our CMDFusion can be tested on the whole LIDAR point cloud. However, our CMDFusion includes both 2D-to-3D and 3D-to-2D fusion while 2DPASS only includes 3D-to-2D fusion, so our method surpasses the 2DPASS according to Table~\ref{tab:kitti_seg}. Note that 2DPASS only released the codebase and the checkpoint without the validation set involved in the training set and instance-level augmentation, so we retrain their model following the same setting and evaluate on the test set. We also try their released checkpoint on the test set and find that both of them achieve a similar mIoU (67.7). We follow the same setting for fair comparison and our method achieves the better performance (68.6 mIoU). We also try the instance-level augmentation from Polarmix~\cite{xiao2022polarmix} on 2DPASS and our method, and our method still surpasses the 2DPASS by 0.6 mIoU. Note that since 2DPASS does not release the code to reproduce the performance reported in their paper, we only compare with them under the same training settings, where our method achieves the better performance. To avoid the mis-correspondence between images and LIDAR point cloud brought by the instance-level augmentation, we do not involve the camera branch during finetuning, and use the frozen 2D knowledge branch to provide 2D information and only finetune the 3D LIDAR branch. In general, our method achieves the best performance among all public methods.

\begin{table*}[t]
\caption{Runtime analysis. $^*$ means accelerated using TensorRT.}
\vspace{-0.5cm}
\begin{center}
\begin{tabular}{l|ccc|cccc|c}
\toprule [1pt]
\multicolumn{1}{c}{}
& \multicolumn{3}{c}{Training} & \multicolumn{4}{c}{Inference}  \\
Method &Modality & \#FLOPs &\#Params. &Modality & \#FLOPs &\#Params. &Speed &mIoU\\ \midrule
PointPainting$^*$~\cite{vora2020pointpainting} &L+C & 51.0 G & 28.1 M &L+C & 51.0 G & 28.1 M & 2.3 ms & - \\
    RGBAL$^*$~\cite{Madawy2019RGBAL} &L+C & 55.0 G & 13.2 M &L+C & 55.0 G & 13.2 M & 2.7 ms & - \\
    SalsaNext$^*$~\cite{cortinhal2020salsanext}&L+C & 31.4 G & 6.7 M &L+C & 31.4 G & 6.7 M & 1.6 ms & 72.2  \\
    Cylinder3D~\cite{cylinder3d} &L &  - & 55.9 M &L &  - & 55.9 M & 62.5 ms & 76.1 \\
    PMF$^*$~\cite{perception} &L+C & 854.7 G & 36.3 M &L+C & 854.7 G & 36.3 M & 22.3 ms & 76.9 \\
    PMF~\cite{perception} &L+C & 854.7 G & 36.3 M &L+C & 854.7 G & 36.3 M & 125.0 ms & 76.9 \\
    2DPASS~\cite{2dpass} &L+C &250.3 G &31.0 M &L &111.3 G &2.36 M &50.0 ms &76.4
    \\
    CMDFusion &L+C &719.6 G &52.2 M &L &373.5 G &7.04 M &125.0 ms &\textbf{77.7} \\
\bottomrule [1pt]
\end{tabular}
\end{center}
\label{tab:time}
\end{table*}

\subsubsection{Results on NuScenes} Table~\ref{tab:nus_seg} shows that our method achieves better performance (2.0 mIoU) than 2DPASS. Similar to the SemanticKITTI, the performance of 2DPASS comes from the higher one between our retrained model and their released checkpoint. Unlike the SemanticKITTI dataset, the NuScenes dataset provides 6 images to cover the FOV of the LIDAR, so the 2D-to-3D fusion methods like PMF~\cite{perception} and 2D3DNet~\cite{genova2021learning} can also be evaluated on the whole LIDAR point cloud. Among all fusion-based methods, our CMDFusion achieves the best performance.

\subsubsection{Visualization} 
\begin{figure}[t]
\centering
\includegraphics[width=3.3in]{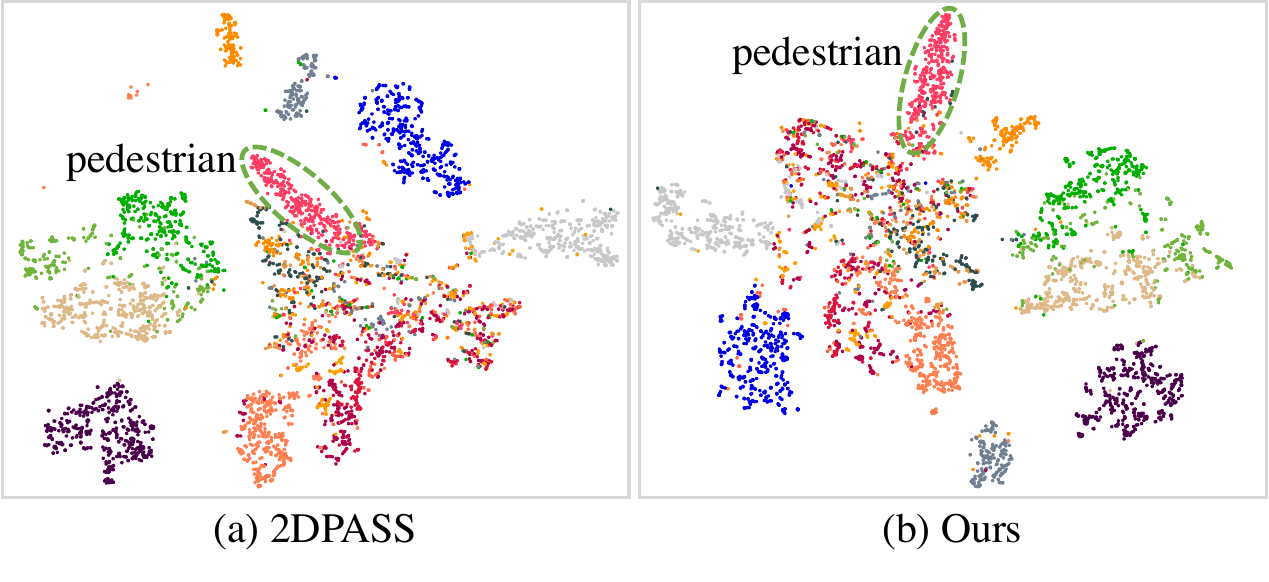}
\vspace{-0.5cm}
\caption{Feature visualization via t-SNE. left: 2DPASS. Right: Ours.}
\label{fig:tsne}
% \vspace{-1.5em} % for figure gap smaller
\end{figure}

We provide two samples from SemanticKITTI and NuScenes datasets in Fig.~\ref{fig:vis}. The top sample shows that 2DPASS and our method have less error on the building compared to the SPVCNN, which illustrates the effectiveness of multi-modality fusion. Besides, our method has better results on the car and truck than 2DPASS, because 2D-to-3D fusion is involved in our method but not in the 2DPASS. In addition, we visualize the feature representation of 2DPASS and our method on the NuScenes dataset. As shown in Fig.~\ref{fig:tsne}, our method has more discriminative features, \textit{e.g.}, the pedestrian class is more separable in our method than 2DPASS.

\subsection{Runtime Analysis} Table~\ref{tab:time} provides the runtime analysis on the NuScenes dataset. PointPainting, RGBAL, and PMF use 2D networks for semantic segmentation since the input is range-view or perspective-view, so they can be accelerated using TensorRT by a large margin (125.0 to 22.3 ms for the PMF method). In contrast, the 3D network in Cylinder3D, 2DPASS, and our method cannot be accelerated by TensorRT. Compared to PMF without TensorRT, our method has a smaller number of FLOPs and parameters during inference, while sharing the same runtime. Compared to 2DPASS, our method achieves better performance since two 3D networks are used during inference (2D Knowledge branch and 3D LIDAR branch), which inevitably consumes more runtime.

\subsection{Ablation Study}
We conduct a careful ablation study to show the effectiveness of different modules in our method. The comprehensive ablation results are based on the Semantic-O dataset since the classical 2D-to-3D fusion without CMD can only be applied on the points in the overlapping FOVs of LIDAR and camera. The results are in Table~\ref{tab:abla}. The baseline refers to a single SPVCNN 3D network. We can see that both 3D-to-2D fusion and 2D-to-3D fusion are helpful, but 2D-to-3D fusion brings more performance gain since the camera information is explicitly injected into the LIDAR branch. After we replace the camera branch (CB) with a frozen CB pre-trained on Cityscapes, the performance is further improved. The reason may be that the pre-trained camera branch could provide additional information for the current LIDAR point cloud dataset. Then we introduce cross-modality distillation (CMD) to let a 3D network output the 2D information so that the model could be trained on the whole dataset rather than the overlapping FOVs of the camera and LIDAR. As a result, the performance is greatly boosted by the CMD. Similar to 2DPASS, we also apply the voting test-time augmentation (TTA), \textit{i.e.}, rotating the input point cloud with 12 angles around the Z axis and averaging the prediction scores as the final outputs. TTA brings better performance by 2.46 mIoU.

\begin{table}[t]
\caption{Ablation study on SemanticKITTI-O (validation set).}
\vspace{-0.5cm}
\begin{center}
\renewcommand\tabcolsep{4pt}
\begin{tabular}{cccccc|c}
\toprule [1pt]
Baseline &3D-to-2D &2D-to-3D &Pretrained CB &CMD &TTA &mIoU\\ \midrule
\checkmark&  &        &     &      & & 58.14   \\
\checkmark&\checkmark &  &  &   & & 58.26 \\
\checkmark&\checkmark & \checkmark &  & &  & 60.42 \\
\checkmark &\checkmark & \checkmark &\checkmark & & &62.03\\
\checkmark &\checkmark & \checkmark &\checkmark &\checkmark & &67.62 \\
\checkmark &\checkmark & \checkmark &\checkmark &\checkmark &\checkmark &70.08 \\
\bottomrule [1pt]
\end{tabular}
\end{center}
\label{tab:abla}
\end{table}

%% file: latex/5_conclu.tex
\section{CONCLUSION}
In this paper, we propose a Bidirectional Fusion Network with Cross-Modality Knowledge Distillation (CMDFusion) to fuse the information of the camera and LIDAR for better LIDAR semantic segmentation. Compared to the 2D-to-3D fusion-based method PMF~\cite{perception}, our proposed Cross-Modality Distillation (CMD) module solves the problem that the camera branch cannot output the 2D information for those points out of the FOV of the camera. Compared to 3D-to-2D fusion-based method 2DPASS~\cite{2dpass}, our proposed Bidirectional Fuision Block (BFB) contains additional 2D-to-3D fusion, which explicitly strengthens the 3D information through 2D information injection for better LIDAR semantic segmentation. We show the effectiveness of our proposed method through comprehensive experiments on SemanticKITTI and NuScenes datasets. Overall, we provide an alternative approach to fully utilize the multi-modality information for 3D semantic segmentation, and introduce a new and feasible way to solve the problem that multi-sensors' FOVs are not overlapping. We hope this paper can provide inspiration for future work in autonomous vehicles and robots.

\section*{ACKNOWLEDGMENT}
This work is supported by Alibaba Group through Alibaba Research Intern Program.

%% file: ms.bbl
\begin{thebibliography}{10}
\providecommand{\url}[1]{#1}
\csname url@samestyle\endcsname
\providecommand{\newblock}{\relax}
\providecommand{\bibinfo}[2]{#2}
\providecommand{\BIBentrySTDinterwordspacing}{\spaceskip=0pt\relax}
\providecommand{\BIBentryALTinterwordstretchfactor}{4}
\providecommand{\BIBentryALTinterwordspacing}{\spaceskip=\fontdimen2\font plus
\BIBentryALTinterwordstretchfactor\fontdimen3\font minus
  \fontdimen4\font\relax}
\providecommand{\BIBforeignlanguage}[2]{{%
\expandafter\ifx\csname l@#1\endcsname\relax
\typeout{** WARNING: IEEEtran.bst: No hyphenation pattern has been}%
\typeout{** loaded for the language `#1'. Using the pattern for}%
\typeout{** the default language instead.}%
\else
\language=\csname l@#1\endcsname
\fi
#2}}
\providecommand{\BIBdecl}{\relax}
\BIBdecl

\bibitem{perception}
Z.~Zhuang, R.~Li, K.~Jia, Q.~Wang, Y.~Li, and M.~Tan, ``Perception-aware
  multi-sensor fusion for 3d lidar semantic segmentation,'' in
  \emph{Proceedings of the IEEE/CVF International Conference on Computer
  Vision}, 2021, pp. 16\,280--16\,290.

\bibitem{2dpass}
X.~Yan, J.~Gao, C.~Zheng, C.~Zheng, R.~Zhang, S.~Cui, and Z.~Li, ``2dpass: 2d
  priors assisted semantic segmentation on lidar point clouds,'' in
  \emph{European Conference on Computer Vision}.\hskip 1em plus 0.5em minus
  0.4em\relax Springer, 2022, pp. 677--695.

\bibitem{spvcnn}
H.~Tang, Z.~Liu, S.~Zhao, Y.~Lin, J.~Lin, H.~Wang, and S.~Han, ``Searching
  efficient 3d architectures with sparse point-voxel convolution,'' in
  \emph{European Conference on Computer Vision (ECCV)}, 2020.

\bibitem{rangenet}
A.~Milioto, I.~Vizzo, J.~Behley, and C.~Stachniss, ``Rangenet++: Fast and
  accurate lidar semantic segmentation,'' in \emph{2019 IEEE/RSJ international
  conference on intelligent robots and systems (IROS)}.\hskip 1em plus 0.5em
  minus 0.4em\relax IEEE, 2019, pp. 4213--4220.

\bibitem{cylinder3d}
H.~Zhou, X.~Zhu, X.~Song, Y.~Ma, Z.~Wang, H.~Li, and D.~Lin, ``Cylinder3d: An
  effective 3d framework for driving-scene lidar semantic segmentation,''
  \emph{arXiv preprint arXiv:2008.01550}, 2020.

\bibitem{rpvnet}
J.~Xu, R.~Zhang, J.~Dou, Y.~Zhu, J.~Sun, and S.~Pu, ``Rpvnet: A deep and
  efficient range-point-voxel fusion network for lidar point cloud
  segmentation,'' in \emph{Proceedings of the IEEE/CVF International Conference
  on Computer Vision}, 2021, pp. 16\,024--16\,033.

\bibitem{semantickitti}
J.~Behley, M.~Garbade, A.~Milioto, J.~Quenzel, S.~Behnke, C.~Stachniss, and
  J.~Gall, ``Semantickitti: A dataset for semantic scene understanding of lidar
  sequences,'' in \emph{Proceedings of the IEEE International Conference on
  Computer Vision}, 2019, pp. 9297--9307.

\bibitem{nuscenes}
H.~Caesar, V.~Bankiti, A.~H. Lang, S.~Vora, V.~E. Liong, Q.~Xu, A.~Krishnan,
  Y.~Pan, G.~Baldan, and O.~Beijbom, ``nuscenes: A multimodal dataset for
  autonomous driving,'' in \emph{CVPR}, 2020.

\bibitem{Randla-Net}
Q.~Hu, B.~Yang, L.~Xie, S.~Rosa, Y.~Guo, Z.~Wang, N.~Trigoni, and A.~Markham,
  ``Learning semantic segmentation of large-scale point clouds with random
  sampling,'' \emph{IEEE Transactions on Pattern Analysis and Machine
  Intelligence}, 2021.

\bibitem{thomas2019kpconv}
H.~Thomas, C.~R. Qi, J.-E. Deschaud, B.~Marcotegui, F.~Goulette, and L.~J.
  Guibas, ``Kpconv: Flexible and deformable convolution for point clouds,'' in
  \emph{ICCV}, 2019.

\bibitem{wu2019pointconv}
W.~Wu, Z.~Qi, and L.~Fuxin, ``Pointconv: Deep convolutional networks on 3d
  point clouds,'' in \emph{CVPR}, 2019.

\bibitem{qi2017pointnet}
C.~R. Qi, H.~Su, K.~Mo, and L.~J. Guibas, ``Pointnet: Deep learning on point
  sets for 3d classification and segmentation,'' in \emph{CVPR}, 2017.

\bibitem{qi2017pointnet++}
C.~R. Qi, L.~Yi, H.~Su, and L.~J. Guibas, ``Pointnet++: Deep hierarchical
  feature learning on point sets in a metric space,'' \emph{arXiv preprint
  arXiv:1706.02413}, 2017.

\bibitem{Minkowski}
C.~Choy, J.~Gwak, and S.~Savarese, ``4d spatio-temporal convnets: Minkowski
  convolutional neural networks,'' in \emph{Proceedings of the IEEE/CVF
  Conference on Computer Vision and Pattern Recognition}, 2019, pp. 3075--3084.

\bibitem{SparseConv}
B.~Graham, M.~Engelcke, and L.~van~der Maaten, ``3d semantic segmentation with
  submanifold sparse convolutional networks,'' in \emph{Proceedings of the IEEE
  Conference on Computer Vision and Pattern Recognition}, 2018, pp. 9224--9232.

\bibitem{polarnet}
Y.~Zhang, Z.~Zhou, P.~David, X.~Yue, Z.~Xi, B.~Gong, and H.~Foroosh,
  ``Polarnet: An improved grid representation for online lidar point clouds
  semantic segmentation,'' in \emph{CVPR}, 2020.

\bibitem{fusionnet}
F.~Zhang, J.~Fang, B.~Wah, and P.~Torr, ``Deep fusionnet for point cloud
  semantic segmentation,'' in \emph{European Conference on Computer
  Vision}.\hskip 1em plus 0.5em minus 0.4em\relax Springer, 2020, pp. 644--663.

\bibitem{xu2021rpvnet}
J.~Xu, R.~Zhang, J.~Dou, Y.~Zhu, J.~Sun, and S.~Pu, ``Rpvnet: A deep and
  efficient range-point-voxel fusion network for lidar point cloud
  segmentation,'' in \emph{Proceedings of the IEEE/CVF International Conference
  on Computer Vision}, 2021, pp. 16\,024--16\,033.

\bibitem{liu2022bevfusion}
Z.~Liu, H.~Tang, A.~Amini, X.~Yang, H.~Mao, D.~Rus, and S.~Han, ``Bevfusion:
  Multi-task multi-sensor fusion with unified bird's-eye view representation,''
  \emph{arXiv preprint arXiv:2205.13542}, 2022.

\bibitem{liang2022bevfusion}
T.~Liang, H.~Xie, K.~Yu, Z.~Xia, Z.~Lin, Y.~Wang, T.~Tang, B.~Wang, and
  Z.~Tang, ``{BEVF}usion: A simple and robust li{DAR}-camera fusion
  framework,'' in \emph{Advances in Neural Information Processing Systems},
  2022.

\bibitem{yang2022deepinteraction}
Z.~Yang, J.~Chen, Z.~Miao, W.~Li, X.~Zhu, and L.~Zhang, ``Deepinteraction: 3d
  object detection via modality interaction,'' in \emph{NeurIPS}, 2022.

\bibitem{he2016deep}
K.~He, X.~Zhang, S.~Ren, and J.~Sun, ``Deep residual learning for image
  recognition,'' in \emph{Proceedings of the IEEE conference on computer vision
  and pattern recognition}, 2016, pp. 770--778.

\bibitem{cordts2016cityscapes}
M.~Cordts, M.~Omran, S.~Ramos, T.~Rehfeld, M.~Enzweiler, R.~Benenson,
  U.~Franke, S.~Roth, and B.~Schiele, ``The cityscapes dataset for semantic
  urban scene understanding,'' in \emph{Proceedings of the IEEE conference on
  computer vision and pattern recognition}, 2016, pp. 3213--3223.

\bibitem{wu2019squeezesegv2}
B.~Wu, X.~Zhou, S.~Zhao, X.~Yue, and K.~Keutzer, ``Squeezesegv2: Improved model
  structure and unsupervised domain adaptation for road-object segmentation
  from a lidar point cloud,'' in \emph{2019 International Conference on
  Robotics and Automation (ICRA)}.\hskip 1em plus 0.5em minus 0.4em\relax IEEE,
  2019, pp. 4376--4382.

\bibitem{xu2020squeezesegv3}
C.~Xu, B.~Wu, Z.~Wang, W.~Zhan, P.~Vajda, K.~Keutzer, and M.~Tomizuka,
  ``Squeezesegv3: Spatially-adaptive convolution for efficient point-cloud
  segmentation,'' in \emph{European Conference on Computer Vision}.\hskip 1em
  plus 0.5em minus 0.4em\relax Springer, 2020, pp. 1--19.

\bibitem{cortinhal2020salsanext}
T.~Cortinhal, G.~Tzelepis, and E.~E. Aksoy, ``Salsanext: Fast semantic
  segmentation of lidar point clouds for autonomous driving,'' \emph{arXiv
  preprint arXiv:2003.03653}, 2020.

\bibitem{choy20194d}
C.~Choy, J.~Gwak, and S.~Savarese, ``4d spatio-temporal convnets: Minkowski
  convolutional neural networks,'' in \emph{IEEE Conference on Computer Vision
  and Pattern Recognition}, 2019, pp. 3075--3084.

\bibitem{vora2020pointpainting}
S.~Vora, A.~H. Lang, B.~Helou, and O.~Beijbom, ``Pointpainting: Sequential
  fusion for 3d object detection,'' in \emph{IEEE Conference on Computer Vision
  and Pattern Recognition}, 2020, pp. 4604--4612.

\bibitem{Madawy2019RGBAL}
K.~E. Madawy, H.~Rashed, A.~E. Sallab, O.~Nasr, H.~Kamel, and S.~Yogamani,
  ``Rgb and lidar fusion based 3d semantic segmentation for autonomous
  driving,'' \emph{IEEE Intelligent Transportation Systems Conference}, pp.
  7--12, 2019.

\bibitem{alonso20203d}
I.~Alonso, L.~Riazuelo, L.~Montesano, and A.~C. Murillo, ``3d-mininet: Learning
  a 2d representation from point clouds for fast and efficient 3d lidar
  semantic segmentation,'' \emph{arXiv preprint arXiv:2002.10893}, 2020.

\bibitem{tatarchenko2018tangent}
M.~Tatarchenko, J.~Park, V.~Koltun, and Q.-Y. Zhou, ``Tangent convolutions for
  dense prediction in 3d,'' in \emph{Proceedings of the IEEE Conference on
  Computer Vision and Pattern Recognition}, 2018, pp. 3887--3896.

\bibitem{yan2020pointasnl}
X.~Yan, C.~Zheng, Z.~Li, S.~Wang, and S.~Cui, ``Pointasnl: Robust point clouds
  processing using nonlocal neural networks with adaptive sampling,'' in
  \emph{Proceedings of the IEEE/CVF Conference on Computer Vision and Pattern
  Recognition}, 2020, pp. 5589--5598.

\bibitem{cheng20212}
R.~Cheng, R.~Razani, E.~Taghavi, E.~Li, and B.~Liu, ``Af2-s3net: Attentive
  feature fusion with adaptive feature selection for sparse semantic
  segmentation network,'' in \emph{Proceedings of the IEEE/CVF conference on
  computer vision and pattern recognition}, 2021, pp. 12\,547--12\,556.

\bibitem{liong2020amvnet}
V.~E. Liong, T.~N.~T. Nguyen, S.~Widjaja, D.~Sharma, and Z.~J. Chong, ``Amvnet:
  Assertion-based multi-view fusion network for lidar semantic segmentation,''
  \emph{arXiv preprint arXiv:2012.04934}, 2020.

\bibitem{genova2021learning}
K.~Genova, X.~Yin, A.~Kundu, C.~Pantofaru, F.~Cole, A.~Sud, B.~Brewington,
  B.~Shucker, and T.~Funkhouser, ``Learning 3d semantic segmentation with only
  2d image supervision,'' in \emph{2021 International Conference on 3D Vision
  (3DV)}.\hskip 1em plus 0.5em minus 0.4em\relax IEEE, 2021, pp. 361--372.

\bibitem{xiao2022polarmix}
A.~Xiao, J.~Huang, D.~Guan, K.~Cui, S.~Lu, and L.~Shao, ``Polarmix: A general
  data augmentation technique for lidar point clouds,'' \emph{arXiv preprint
  arXiv:2208.00223}, 2022.

\end{thebibliography}
